\documentclass[10pt, a4paper]{article}

\usepackage{lrec-coling2024} 
\usepackage{verbatim}
\usepackage{algorithm}
\usepackage{algpseudocode}
\usepackage{caption}
\usepackage{float}
\usepackage{fullwidth}

\DeclareCaptionFormat{algor}{%
  \hrulefill\par\offinterlineskip\vskip1pt%
    \textbf{#1#2}#3\offinterlineskip\hrulefill}
\DeclareCaptionStyle{algori}{singlelinecheck=off,format=algor,labelsep=space}
\usepackage[font={small}]{caption}
\usepackage[size=scriptsize]{todonotes}
\usepackage{soul}

\newcommand{\para}[1]{\vspace{-0.2cm}\paragraph{#1}}

\title{FastSpell: the LangId Magic Spell}

\name{
Marta Bañón, Jaume Zaragoza-Bernabeu, \\ {\bf  \large Gema Ramírez-Sánchez, Sergio Ortiz-Rojas
}}

\address{
Prompsit Language Engineering, S.L., Spain \\
\{mbanon, jzaragoza, gramirez, sortiz\}@prompsit.com \\
}

\abstract{
Language identification is a crucial component in the automated production of language resources, particularly in multilingual and big data contexts. However, commonly used language identifiers struggle to differentiate between similar or closely-related languages. This paper introduces FastSpell, a language identifier that combines fastText (a pre-trained language identifier tool) and Hunspell (a spell checker) with the aim of having a refined second-opinion before deciding which language should be assigned to a text. We provide a description of the FastSpell algorithm along with an explanation on how to use and configure it. To that end, we motivate the need of such a tool and present a benchmark including some popular language identifiers evaluated during the development of FastSpell. We show how FastSpell is useful not only to improve identification of similar languages, but also to identify new ones ignored by other tools. 
 \\ \newline }

\begin{document}
\maketitleabstract

\section{Introduction}

Language identification is at the core of many NLP pipelines and an essential component in the automatic production of language resources. A bad choice of the technology used to perform language identification may have a crucial impact on the rest of the pipeline and the final results, especially in big data and multilingual data contexts. In such contexts, a complex variety of languages need to be language-identified at scale and three factors are of great importance: number of languages covered, accuracy and speed. However, the choice of language identifiers is mostly based on language coverage as this is usually a crucial factor in the decision. Other factors may be disregarded and, once the choice is made, the rest of the pipeline will need to cope with the possible mistakes made by the selected language identifier. 

In this paper, we present FastSpell, a language identifier that reviews and complements prior language identification processes. It seeks for a compromise between speed and quality, with special focus on similar languages and language varieties. FastSpell requires a language to focus on, which we define as the \texttt{targeted language}, usually coming from a previous language identifier. FastSpell double-checks that targeted language, paying special attention to languages that are often confused with it. To that end, FastSpell first asks fastText\footnote{\url{https://fasttext.cc/}} to give a prediction. Then, only if the language predicted by fastText falls into a group of languages similar to the targeted language, FastSpell refines its decision by performing extra checks with the Hunspell\footnote{\url{http://hunspell.github.io/}} spell-checker. This allows to double-check the prediction made by fastText as well as to discriminate better between similar languages, identify new languages or group them.

FastSpell, distributed under the GPLv3 license,\footnote{\url{https://github.com/mbanon/fastspell/blob/main/LICENSE}} is officially supported and maintained inside the Bitextor\footnote{\url{https://github.com/bitextor}} and Monotextor\footnote{\url{https://github.com/bitextor/monotextor}} pipelines, two free/open-source tools used to produce parallel and monolingual corpora from web-crawled content. It runs in the corpus cleaning phase of these two tools to refine the decisions made by any language identifier applied earlier. It has been used to produce corpora in projects such as ParaCrawl,\footnote{\url{https://paracrawl.eu}} MaCoCu,\footnote{\url{https://macocu.eu}} and HPLT,\footnote{\url{https://hplt-project.org}} all complex in the number and variety of languages and amounts of data processed.

In the following sections, we explain the motivation of building FastSpell (section \ref{section:why}) and why it relies on fastText (section \ref{section:benchmark}), which was chosen after a careful evaluation of available tools. This evaluation, designed and maintained up to date as an independent benchmark, focuses on particularly challenging cases, that is, deciding between similar languages, and reports both accuracy and speed performance. Section \ref{section:formula} describes the FastSpell algorithm while section \ref{section:use} explains how to use and configure it. Finally, section \ref{section:future} draws some conclusions and future working plans.

\section{Why FastSpell?}
\label{section:why}

FastSpell was initially developed as part of the code of the ParaCrawl series of projects aiming at deriving parallel data from web-crawled content \citep{banon-etal-2020-paracrawl}. Language identification of web-crawled data, usually very noisy \citep{quality-at-glance}, is a necessary step to derive language-specific textual corpora from them. After manual inspections of the parallel sentences produced in ParaCrawl, we found several issues with CLD2,\footnote{\url{https://github.com/CLD2Owners/cld2}} the tool used at the moment to identify language at document level, then transferred to sentence level: 
\begin{itemize}
    \item Closely related languages often get mixed up. Especially if one of them has significantly more resources, for example, Spanish and Galician or the Bokmål and Nynorsk variants of Norwegian.
    \item Text containing all or mostly uppercase letters is very often classified as the highest-resourced language using a particular writing system. Many Latin script languages end up badly identified as English, Spanish or French and Cyrillic script languages as Russian.
    \item Languages that use two or more scripts are usually identified with just one of them. For example, although Serbo-Croatian languages can technically be written in Cyrillic and Latin, Cyrillic gets most of the time classified as Serbian and Latin as Croatian. While these are the mostly used scripts for those languages, this just does not cover all cases well enough, e.g. Serbian written in Latin.
\end{itemize}

FastSpell was developed to be able to cope with these issues and refine the decisions made by CLD2 at the beginning of the pipeline. 


\section{Benchmarking Language Identifiers}
\label{section:benchmark}
FastSpell starts by launching automatic language identification over a text, sentences in our case. To be able to make an informed decision on which tool was better suited for this task, we benchmarked  several language identification tools (see Table \ref{fig:F1time}) focusing on performance (runtime) and accuracy (F-score). This benchmark is constantly evolving to incorporate new tools and languages and is publicly available at \url{https://github.com/mbanon/benchmarks} with results available at \url{https://tinyurl.com/2u48kycz}.

The benchmark includes a diverse set of languages added in two batches according to projects needs. The first batch, introduced during the ParaCrawl project, covered Spanish (es), Galician (gl), Catalan (ca), Danish (da), Norwegian Bokmål (nb) and Norwegian Nynorsk (nn). The second one, introduced during the MaCoCu project, included Bulgarian (bg), Czech (cz), Greek (el), Macedonian (mk), Romanian (ro), Slovak (sk), Slovene (sl), Albanian (sq), Maltese (mt), Turkish (tr), Bosnian (bs), Montenegrin (me), Croatian (hr), and Serbian (sr). For convenience, we grouped several times Croatian, Bosnian, Serbian and Montenegrin under the Serbo-Croatian (hbs) macrolanguage. 

\begin{table}[!htb]
    \centering
    \setlength{\tabcolsep}{2pt}
    \resizebox{\columnwidth}{!} {
    \begin{tabular}{|c||ccccc|}
    \hline
     \textbf{} & 
     \textbf{} & \textbf{} & \textbf{F1 scores} & \textbf{} & \textbf{} \\

        \textbf{Lang} & \textbf{pyCLD2} & \textbf{pyCLD3} & \textbf{fastText 176} & \textbf{HeLI} & \textbf{FastSpell} \\
        \hline
        \hline
        \textbf{es} & 0,890	 & 0,932&0,929&\textbf{0,957}&0,954 \\ \hline
        \textbf{gl} & 0,866&0,903&0,808&\textbf{0,939}&0,800\\ \hline
        \textbf{ca} & 0,904&0,935&0,951&\textbf{0,961}&0,935 \\ \hline
        \textbf{da} & 0,914&0,868&0,824&\textbf{0,930}&0,799\\ \hline
        \textbf{nb} & 0,713&-&-&\textbf{0,724}&0,675 \\ \hline
        \textbf{nn} & 0,781&-&0,433&0,787&\textbf{0,810}  \\  \hline
        \textbf{bg} & 0,939&0,959&0,979&0,972&\textbf{0,990}  \\ \hline        
        \textbf{cs} & \textbf{0,965}&0,908&0,918&0,941&0,962 \\ \hline
        \textbf{el} & 0,976&0,964&\textbf{1,000}&0,991&\textbf{1,000} \\ 
        \hline                
        \textbf{mk} & 0,883&0,981&\textbf{0,993}&0,982&0,985 \\ \hline
        \textbf{ro} & 0,960&0,947&0,970&\textbf{0,988}&0,975 \\ \hline
        \textbf{sk} & \textbf{0,966}&0,934&0,917&0,952&0,937 \\ \hline
        \textbf{sl} & 0,925&0,887&0,847&\textbf{0,942}&0,880 \\ \hline
        \textbf{sq} & 0,981&0,986&\textbf{0,993}&0,992&0,990  \\ \hline
        \textbf{mt} & 0,984&0,976&0,896&\textbf{0,996}&0,914\\ \hline 
         \textbf{tr} & 0,978&0,985&\textbf{0,990}&0,983&0,988\\ \hline
        \textbf{bs} & \textbf{0,443}&0,336&0,416&-&0,370   \\ \hline
        \textbf{me} & -&-&-&-&\textbf{0,458}  \\ \hline
        \textbf{hr} &  \textbf{0,641}&0,423&0,557&-&0,541 \\ \hline
        \textbf{sr} &  0,529&0,327&\textbf{0,565}&-&0,493 \\ \hline        
        \textbf{hbs} & 0,940&0,941&0,922&0,982&\textbf{0,983}  \\
         \hline\hline
        \textbf{Avg. Runtime} & \textbf{0,011}&0,097&0,019&2,688&1,076 \\ \hline
        
    \end{tabular}
    }
    \caption{F1 scores and average runtime for all the languages and some of the language identification tools benchmarked. Best scores in bold.}
    \label{fig:F1time}    
\end{table}

For each language in each batch, we built a gold-standard corpus made of sentences. We used SETIMES\footnote{\url{https://opus.nlpl.eu/SETIMES.php}} for Bosnian, Macedonian, Albanian, Serbian and Turkish, MontenegrinSubs \citeplanguageresource{inproceedings} and texts from the Government of Montenegro webpage\footnote{\url{https://www.gov.me/}} for Montenegrin and Paracrawl human annotations \citep{ramirez-sanchez-etal-2022-human} for the other languages. We also built an "anti-gold standard" by combining all sentences in a batch and excluding the sentences for the targeted language (for example, the anti-gold standard for Norwegian Nynorsk included the sentences in the Danish and Norwegian Bokmål gold-standards).

The benchmarked tools and versions are pyCLD2,\footnote{\url{https://github.com/aboSamoor/pycld2}} pyCLD3,\footnote{\url{https://github.com/bsolomon1124/pycld3}} 
langid,\footnote{\url{https://github.com/saffsd/langid.py}} FastLang,\footnote{\url{https://spacy.io/universe/project/spacy_fastlang}} langdetect,\footnote{\url{https://github.com/Mimino666/langdetect}} NLTK,\footnote{\url{https://www.nltk.org/api/nltk.classify.html\#nltk.classify.textcat.TextCat.guess_language}} GuessLanguage,\footnote{\url{https://pypi.org/project/guess_language-spirit}} FastText (lid.176 model compressed and binarized version) and HeLI-OTS \citep{jauhiainen-etal-2022-heli}. The same benchmarking was performed combining Hunspell with PyCLD2 and PyCLD3 instead of fastText.

 As shown in Table \ref{fig:F1time}, in terms of F-scores and runtime, CLD2 would have been the best candidate for language identification in FastSpell but, since it was already used in previous steps in the Bitextor production pipeline, 
we decided to use a different and complementary tool. HeLIOTS would have also been a good choice, but it was, at the moment, up to 100 times slower than fastText on average. For these reasons, fastText was selected as the language identification tool to be included in FastSpell. 

\begin{figure}[!hpt]
    \centering
    \includegraphics[width=1\linewidth]{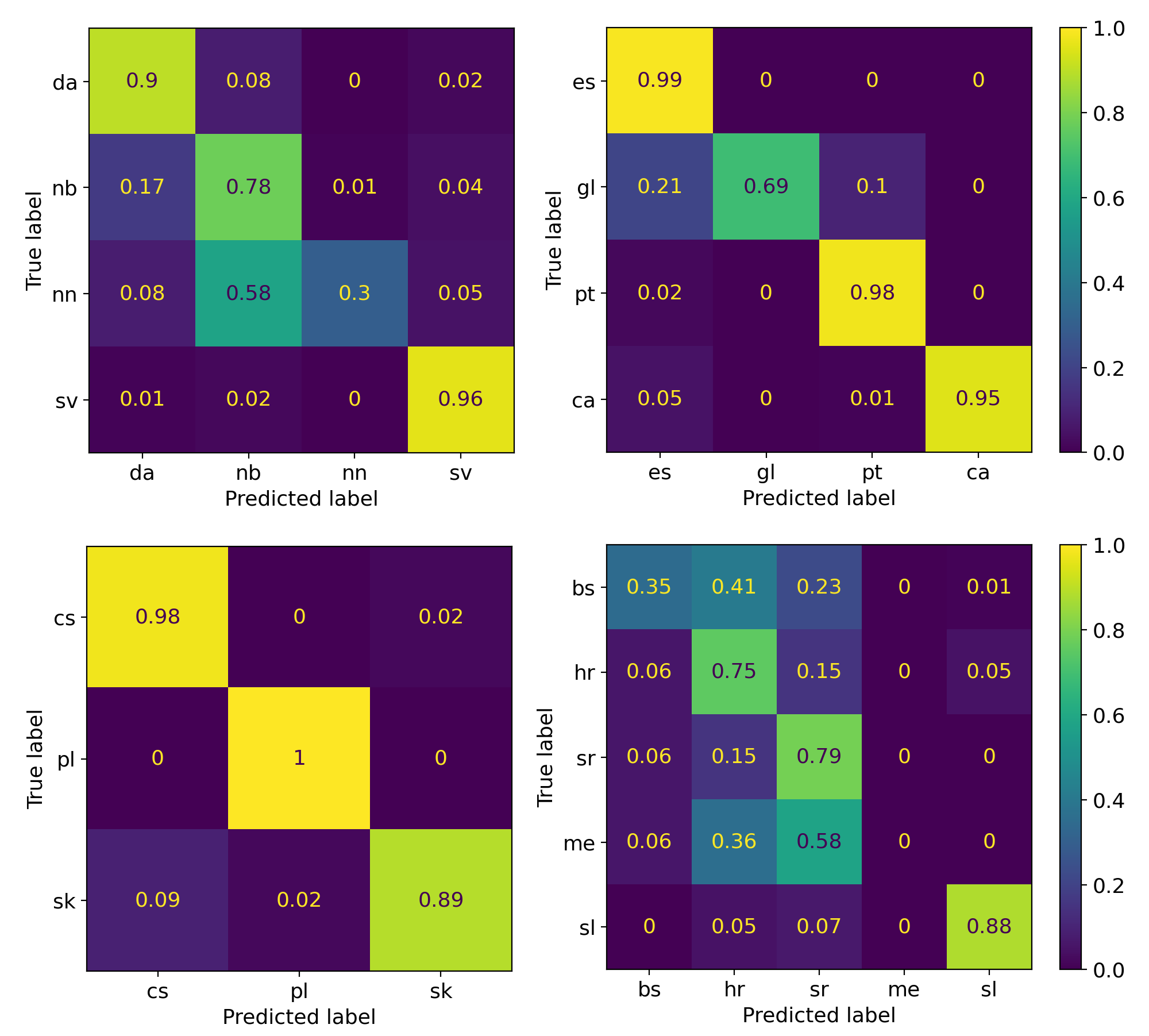}\hfill
    \caption{fastText confusion matrices for some groups of similar languages.}
    \label{fig:confmatrix}
\end{figure}

\section{The FastSpell Spell}
\label{section:formula}
In order to address the issues mentioned in section \ref{section:why} with similar languages being confused, FastSpell makes a two-step decision using two different, well-known tools: \emph{fastText} and \emph{Hunspell}.

FastText~\citep{joulin2016fasttext, joulin2016bag} is a free/open-source library for fast text representation and classification, developed by Meta.\footnote{\url{https://opensource.fb.com}} In particular, the language identification model used in FastSpell is the  \texttt{lid.176.bin} model, trained on data from Wikipedia, Tatoeba and SETIMES and able to recognize up to 176 languages.\footnote{\url{https://fasttext.cc/docs/en/language-identification.html}}

Using fastText alone to make this second prediction was deemed insufficient to make good predictions in many cases. As shown in Figure \ref{fig:confmatrix}, fastText has issues discriminating among similar languages which it confuses frequently (for example, Galician with Spanish), struggling on Norwegian Nynorsk with Norwegian Bokmål, or making inaccurate predictions for many South-Slavic closely-related languages. In other cases, predictions look quite good, but still not completely accurate as in the case of Slovak. Thus, a second step relying on spell-checking results was introduced.     

The technology chosen to perform spell-checking is Hunspell, the well-known free/open-source spell checker and morphological analyser widely used in also well-known tools, such as Mozilla Firefox, Google Chrome or the LibreOffice suite. For spell checking, Hunspell uses affix files and dictionaries, that can be obtained from different sources\footnote{\url{https://github.com/LibreOffice/dictionaries}, \url{https://github.com/wooorm/dictionaries/}, \url{https://extensions.libreoffice.org/?q=spellchecker}, \url{http://hlt.sztaki.hu/resources/hunspell/}} independently or be custom-built by users.

FastSpell focuses on a given language (the targeted language), that is provided as a parameter. Given a text (usually, a sentence), FastSpell will first predict its language by using fastText. For efficiency, only if the predicted language is the targeted language or a similar language according to a configurable list (see section \ref{fig:similars}), FastSpell will try to refine the fastText prediction by checking the sentence spelling with Hunspell for the targeted language and its similar languages. Depending on the ratio of spelling errors for each of the spell-checked languages, FastSpell will confirm the targeted language as the winner or replace it by the similar language with the lowest number of spelling errors. This is true for the \textit{aggressive} mode 
. However, in the \textit{conservative} mode, when there is not a clear winner after the spell-checking step, the language for that text can be tagged as "unknown". The full algorithm is shown in Appendix \ref{appendix:fullalgorithm}. 

\section{Using FastSpell}
\label{section:use}

FastSpell can be installed to be used as a CLI tool or as a Python package. In both cases, besides a text, the only parameters that are needed are the targeted language (i.e. the language predicted by a prior language identification tool and the mode (either aggressive or conservative). In the example below, FastSpell receives English (\texttt{en}) as the targeted language and conservative (\texttt{cons}) as the mode. For the first sentence (\texttt{Hello, world}), FastSpell outputs English as the predicted language, for the second one (\texttt{Hola, mundo}), it outputs Spanish (\texttt{es}), refining the a priori prediction.

\begin{footnotesize}
\begin{verbatim}
from fastspell import FastSpell
fsobj=FastSpell("en", mode="cons")
fsobj.getlang("Hello, world")
#'en'
fsobj.getlang("Hola, mundo")
#'es'
\end{verbatim}
\end{footnotesize}

\para{Out-of-the-box} FastSpell comes pre-configured for several targeted languages and the necessary linguistic resources (fastText model and Hunspell dictionaries) are installed as a dependency.\footnote{\url{https://pypi.org/project/fastspell-dictionaries/}} 
This default configuration mainly focuses on the languages of interest of the three projects in which FastSpell has been used: Paracrawl \citep{banon-etal-2020-paracrawl}, MaCoCu \citep{non-etal-2022-macocu} and HPLT \citep{aulamo-etal-2023-hplt}. Besides addressing the issue of identifying similar or closely related languages (for example, Spanish and Galician), we also provide a solution to identify single languages from a macrolanguage (languages in the Serbo-Croatian family or Norwegian Bokmål/Nynorsk), languages not supported by the current fastText model (for example, Montenegrin) and languages that in principle are not similar but that fastText seems to confuse (for example, Somali and English). The pre-configured targeted languages and the ones similar to each of them are shown in  Appendix \ref{appendix:similarlanguages}.

\para{Custom configuration} The default configuration of FastSpell might not be suitable in some cases, for example, when it does not support the targeted language by default, when there's a need to change the pre-defined similar languages for a targeted language, or when a different Hunspell dictionary needs to be used. FastSpell is easily customizable in these cases, only requiring to modify some configuration text files located in the fastspell/config directory.

One of those files is \texttt{similar.yaml} (whose first lines can be seen in Figure \ref{fig:similars}), a  file containing the pre-defined targeted languages and their associated similar languages. These languages (see the full list in Appendix \ref{appendix:similarlanguages}) will be double-checked with Hunspell after being predicted by fastText. Adding  or removing a new language from this file will activate or deactivate  the spellchecking step in case it matches the criteria of the FastSpell algorithm (see section \ref{section:formula}). The list of similar languages associated to a targeted language can also be modified. Note that the list of similar languages is not necessarily symmetrical: a targeted language \textit{A} may have language \textit{B} as similar, but a targeted language \textit{B} may not have language \textit{A} as similar instead. Indeed, it may not even be necessary to have language \textit{B} configured as a targeted language. 

\begin{figure}
\begin{footnotesize}
\begin{verbatim}
#Targeted langs(keys) dict for
#mistakeable languages (values)
similar:
    af: [nl, de, af]
    az: [tr, az]
    be: [ru, uk, be]
    bg: [mk, ru, bg]
    bs: [hr, sr, sl, bs]
    ca: [es, oc, ca]
    cs: [sk, cs]
    cy: [ga, en, cy]    
\end{verbatim}
\caption{First lines of the default \textit{similar.yaml} file}
\label{fig:similars}
\end{footnotesize}
\end{figure}

The \texttt{hunspell.yaml} file, which contains Hunspell-related information such as the path to the location of Hunspell dictionaries and the name of the dictionary for each language, may also be configured. This allows some flexibility in usage, for example, to assemble a dictionary for a macrolanguage and use it as a targeted or similar language. This is the case of Serbo-Croatian (hbs), for which a single Hunspell dictionary gathers together Serbian, Croatian and Bosnian in FastSpell.  

FastSpell results, included in \ref{fig:F1time}, show how some languages not supported or badly supported by other tools can be more reliably identified using FastSpell, as is the case of Montenegrin or Norwegian Nynorsk. 


\section{Conclusions and Future Work}
\label{section:future}
We have presented FastSpell, a second-opinion language identifier that reviews and refines decisions made by a previous language identifier from which a targeted language is set. FastSpell is able to distinguish better between closely-related languages or to discover new languages or language varieties  not predicted by a language identifier (for example, Norwegian Nynorsk), usually hidden or confused with a larger-resource language (for example, Norwegian Bokmål). There is still room for enhancements that could be added to FastSpell, for example:

\begin{itemize}
    \item Exploring possible replacements for the current fastText model, lid.176.bin, by, for example, the 201-language model introduced in \citep{burchell-etal-2023-open}.
    \item Exploring faster implementations of fastText such as fasterText.\footnote{\url{https://github.com/kpu/fasterText}} 
    \item Curating Hunspell dictionaries for languages not having publicly available dictionaries to extend FastSpell's language support. For example, Pashto seems to be frequently miss-labelled as Arabic, but we have not been able to find any Pashto dictionaries that can be integrated into FastSpell. A similar situation is found with Sindhi and Farsi.
    \item Exploring proper tokenization and/or stemming of sentences before spellchecking with Hunspell to improve language identification accuracy after it. Since Hunspell is based in dictionaries of known words, spellchecking stems instead of full words will probably result in less false negatives, specially in inflected languages. 
    \item Adding non-targeted identification, that is, not focusing on a given language (by applying extra checks only for it and its similar languages), but applying the extra checks for any identified language. This is expected to be substantially slower, but will provide more accurate language identification.    
    \item Exploring using different error thresholds, depending on the targeted language.
    \item Write a Hunspell-like engine which is capable to process more than one language at once to avoid repeated checks.  
\end{itemize}

Current language extensions and enhancements to FastSpell have been made inside the HPLT project, still ongoing, which has recently produced language resources for more than 75 languages \citeplanguageresource{degibert2024new}. Among those, there are many that have different variants or are closely-related languages which have benefited from FastSpell refinements. 


\para{Acknowledgements}
This work has been supported by the three ParaCrawl projects (paracrawl.eu) funded by the Connecting Europe Facility of the European Union 2014--2020 (CEF Telecom) and an additional project, MaCoCu (macocu.eu), also funded by the same programme under Grant Agreement No. INEA/CEF/ICT/A2020/2278341, all already finished. It is now being supported by the European Union’s Horizon Europe research and innovation programme under grant agreement No 101070350 and from UK Research and Innovation (UKRI) under the UK government’s Horizon Europe funding guarantee [grant number 10052546] through the HPLT project (hplt-project.eu.org). We thank professor Mikel L. Forcada for its thorough review and contributions to this paper.

\section{Bibliographical References}
\label{sec:reference}
\vspace{-0.8cm} 
\bibliographystyle{lrec-coling2024-natbib}
\bibliography{references}


\section{Language Resource References}
\label{lr:ref}
\vspace{-0.8cm} 
\bibliographystylelanguageresource{lrec-coling2024-natbib}
\bibliographylanguageresource{languageresource}

\clearpage
\clearpage
\appendix
\section{Pre-configured similar languages}
\label{appendix:similarlanguages}
\begin{table}[h!]
\centering
\begin{tabular}{|c|c|}
\hline
\textbf{Targeted language} & \textbf{Similar languages} \\
\hline
\hline
Afrikaans & Dutch, German \\
\hline
Azerbaijani & Turkish \\
\hline
Belarusian & Russian, Ukrainian \\
\hline
Bulgarian & Macedonian, Russian \\
\hline
Bosnian & Croatia, Serbian, Slovene \\
\hline
Catalan & Spanish, Occitan \\
\hline
Czech & Slovak \\
\hline
Welsh & Irish, English \\
\hline
Danish & Norwegian Bokmål, Swedish \\
\hline
Spanish & Galician, Catalan \\
\hline
Farsi & Arabic, Azerbaijani \\
\hline
Irish & Welsh, English \\
\hline
Galician & Spanish, Portuguese \\
\hline
Serbo-Croatian (Latin) & Slovene \\
\hline
Serbo-Croatian (Cyrillic) & Russian, Macedonian, Bulgarian \\
\hline
Hindi & Marathi, Nepali \\
\hline
Croatian & Bosnian, Serbian, Slovene \\
\hline
Indonesian & Malay \\
\hline
Icelandic & Danish, Norwegian Bokmål, Norwegian Nynorsk, Swedish \\
\hline
Hebrew & Yiddish \\
\hline
Kazakh & Kyrgyz, Tatar, Russian \\
\hline
Kyrgyz & Russian, Kazakh, Tatar, Mongolian \\
\hline
Latvian & Lithuanian \\
\hline
Montenegrin & Croatian, Serbian, Slovene, Bosnian \\
\hline
Macedonian & Bulgarian, Serbian, Russian \\
\hline
Mongolian  & Russian, Kyrgyz, Bulgarian \\
\hline
Marathi & Hindi \\
\hline
Malay & Indonesian \\
\hline
Norwegian Bokmål & Danish, Swedish, Norwegian Nynorsk \\
\hline
Nepali & Marathi, Hindi \\
\hline
Dutch & Afrikaans \\
\hline 
Norwegian Nynorsk & Norwegian Bokmål, Danish, Swedish \\
\hline
Norwegian & Danish, Swedish, Norwegian Nynorsk \\
\hline
Portuguese & Spanish, Galician \\
\hline
Russian & Ukrainian, Bulgarian \\
\hline
Slovak & Czech, Polish \\
\hline
Slovene & Serbian, Croatian, Bosnian \\
\hline
Somali & English, Finnish, Welsh, Kannada \\
\hline
Serbian & Bosnian, Croatian, Slovene, Montenegrin \\
\hline
Swedish & Danish, Norwegian Bokmål \\
\hline
Tatar & Kazakh, Kyrgyz, Russian \\
\hline
Ukrainian & Belarusian, Russian, Macedonian, Bulgarian \\
\hline
Urdu   & Farsi, Arabic \\
\hline
Uzbek & Turkish \\
\hline
Yiddish & Hebrew \\
\hline 
\end{tabular}
\caption{FastSpell preconfigured similar languages}
\label{table:fastspell-default-langs}
\end{table}

\clearpage
\begin{figure}[htb]
\section{The FastSpell Algorithm}
\label{appendix:fullalgorithm}
\centering
\begin{minipage}{2\linewidth}
\begin{algorithm}[H]
\begin{algorithmic}
\Function{getLanguage}{target\_lang, sentence, strategy}
\State similar\_langs $\gets$ SimilarLanguages(target\_lang) $\cup$ \{target\_lang\}
\State pred\_FT $\gets$ FastText(lowercase(sentence))
\If{|similar\_langs|=1} 
\Comment{similar\_langs=\{target\_lang\} only}
    \State \Return pred\_FT              
\EndIf         
\If{pred\_FT $\notin$ similar\_langs} 
    \State \Return pred\_FT
\EndIf 
\State candidate\_langs $\gets$ $\emptyset$
\State best\_error\_rate $\gets$ 0
\ForAll{sim\_lang $\in$ similar\_langs} 
    \State relevant\_tokens $\gets$ remove\_uppercased(remove\_non\_alphabetic(tokens(sentence)))
    \State  correct\_tokens $\gets$ collect\_correct\_tokens(Hunspell(relevant\_tokens, sim\_lang))
    \State error\_rate $\gets$ $1-(|$correct\_tokens$|/|$relevant\_tokens$|)$
    \If {error\_rate $\le$ error\_threshold}
        \State candidate\_langs $\gets$ candidate\_langs $\cup$ \{sim\_lang\}
        \If{error\_rate $<$ best\_error\_rate }
             \State best\_error\_rate $\gets$ error\_rate
             \EndIf
    \EndIf
\EndFor 
\State refined\_candidates $\gets$ candidates\_with\_lowest\_error\_rate(candidate\_langs) 
\If { |refined\_candidates| = 1 }
    \State \Return first(refined\_candidates) \Comment{The first and only language in the set}
\ElsIf { |refined\_candidates| > 1}
    \If {strategy = aggressive}
        \If {target\_lang in refined\_candidates} 
            \State \Return target\_lang
        \ElsIf {pred\_FT in refined\_candidates}
	    \State \Return pred\_FT
        \Else 
            \State \Return first(refined\_candidates)  \Comment{The first language will do in case of a draw}
        \EndIf  
    \ElsIf {strategy = conservative}
        \If {target\_lang $\in$ refined\_candidates $\land$ best\_error\_rate = 0} 
            \State \Return target\_lang
	\Else
	    \State \Return unknown\_lang  \Comment{A special code}
        \EndIf  
    \EndIf
\ElsIf{ |refined\_candidates|=0 $\lor$ |candidate\_langs|=0}
    \If  {strategy = aggressive}	
	\State \Return pred\_FT
    \ElsIf {strategy = conservative}
	  \State \Return unknown\_lang
    \EndIf 
\EndIf 
    
\EndFunction
\end{algorithmic}
\end{algorithm}
\end{minipage}
\caption{The FastSpell Algorithm}\label{alg:cap}
\end{figure}

\end{document}